\documentclass[review]{elsarticle}

\usepackage{lineno,hyperref}
\modulolinenumbers[5]

\journal{Journal of \LaTeX\ Templates}

\usepackage{amsmath}
\usepackage{tabularx}
\usepackage{graphicx}
\usepackage{amssymb}
\usepackage[misc]{ifsym}
\usepackage{pifont}
\usepackage{color}
\usepackage{multirow}
\usepackage{booktabs}
\usepackage{threeparttable}

\begin{document}
	
	\begin{frontmatter}
		
		\title{Unified Chinese License Plate Detection and Recognition with High Efficiency}
		
		
		\author{Yanxiang~Gong, Linjie~Deng, Shuai~Tao, Xinchen~Lu, Peicheng~Wu, Zhiwei~Xie, Zheng~Ma}
		
		\author{Mei~Xie\corref{mycorrespondingauthor}}
		\cortext[mycorrespondingauthor]{Corresponding author}
		\ead{mxie@uestc.edu.cn}
		
		\address{School of Information and Communication Engineering, \\University of Electronic Science and Technology of China}
		\begin{abstract}
			Recently, deep learning-based methods have reached an excellent performance on License Plate (LP) detection and recognition tasks. However, it is still challenging to build a robust model for Chinese LPs since there are not enough large and representative datasets. In this work, we propose a new dataset named Chinese Road Plate Dataset (CRPD) that contains multi-objective Chinese LP images as a supplement to the existing public benchmarks. The images are mainly captured with electronic monitoring systems with detailed annotations. To our knowledge, CRPD is the largest public multi-objective Chinese LP dataset with annotations of vertices. With CRPD, a unified detection and recognition network with high efficiency is presented as the baseline. The network is end-to-end trainable with totally real-time inference efficiency (30 fps with 640p). The experiments on several public benchmarks demonstrate that our method has reached competitive performance. The code and dataset will be publicly available at https://github.com/yxgong0/CRPD.
		\end{abstract}
		
		\begin{keyword}
			Chinese license plate dataset; License plate detection and recognition; End-to-end; Real-time
		\end{keyword}
		
	\end{frontmatter}
	
	% \linenumbers
	
	\section{Introduction}
	
	\label{sec:introduction}
	License Plate (LP) detection and recognition are the key parts of intelligent transportation systems because it is the unique identification of vehicles. The relevant methods are widely used on electronic toll payment, parking managing, and traffic monitoring systems. In order to achieve effective detection and recognition, researchers proposed a variety of techniques that are capable of handling the task in most conditions. Before the deep learning era, most of the methods were on the basis of artificial designs~\cite{tradm1,tradm2,alpr}. They utilized hand-crafted features such as colors, shadows and textures, and integrated them by a cascaded strategy with license plate detection, segmentation, and recognition. Although they reached promising performance, the robustness may not be enough for some uncontrolled circumstances like weather, illumination, and rotation since the scheme relies on manually designed features. Then with the explosion of deep learning methods in recent years, most researchers turned their attention to this framework that is able to learn features automatically. The convenient and efficient technique quickly became popular, and networks for detection and recognition also sprung up~\cite{pvw, vlpd, lprec1}. In these works, great successes have been achieved. However, for Chinese LPs, the problem of data scarcity gradually emerges. Existing annotated Chinese LP data that is representative of most scenarios cannot meet the huge demands. Thus, to alleviate the issue, we present our Chinese Road Plate Dataset (CRPD).
	
	Admittedly, there are already some excellent public datasets with LPs~\cite{pvw,reid, caltech,zemris,ccpd,aolp,ssig,ufpr,pku,clpd}. These public benchmarks lay the foundation of various LP processing methods, and our CRPD is an effective supplement to existing Chinese LP datasets, which is more challenging. Images of CRPD are collected from electronic monitoring systems in most provinces of mainland China in different periods and weather conditions. The images contain cars with different statuses and types, and quite a part of the data contains more than one LP in one image. Each image has annotations of (i) LP content. (ii) Locations of four vertices. (iii) LP type. More details will be introduced in Section~\ref{sec:1}.
	
	% ReId~\cite{reid} is a dataset for license plate recognition with 76k images gathered from surveillance cameras on highway toll gates. Caltech~\cite{caltech} and Zemris~\cite{zemris} collected over 600 images from the road and freeways with high-resolution cameras. CCPD~\cite{ccpd} is a large and comprehensive dataset with about 290k images that contain plates with various angles, distances, and illuminations.
	
	As for detection and recognition tasks, most prestigious methods designed the two branches separately. Zhou \textsl{et al.}~\cite{pvw} proposed a scheme for LP detection with Principal Visual Word (PVW) generation and applied bag-of-words in partial-duplicate image search. Chen \textsl{et al.}~\cite{vlpd} put forward a method to detect the vehicles and the LPs simultaneously, where the results can be used for further recognition. The two-stage framework is effective, but the error accumulation problems hinder further progresses. Therefore, end-to-end frameworks are increasingly prevalent. In this paper, we propose an end-to-end trainable Chinese LP detection and recognition network with both high efficiency and satisfactory performance as the baseline of our CRPD. Our method is a unified network that consists of two branches. The branch for detection is based on STELA~\cite{stela}, which is a learned anchor-based detector. It only associates one reference box at each spatial position, which highly reduces the computation to reach a fast running speed. In the recognition branch, we abandon the recognition by segmentation pipeline and utilize a sequence-to-sequence method to accomplish the recognition task. The region-wise features are extracted by the RRoIAlign~\cite{fots} operator and then will be fed into components for recognition. This scheme blurs the line between detection and recognition, which strongly alleviates the error accumulation issues. 
	
	% And a two-stage network was presented in~\cite{twostage} to achieve the detection and recognition of LPs
	
	In summary, there are three main contributions in this work:
	
	\begin{itemize}
		\item We publish a new Chinese LP dataset with more than 30k images, which covers more scenes and administrative regions of mainland China. We argue that this new dataset is more difficult than existing datasets and is also a supplement to the Chinese LP research field.
		\item We propose an end-to-end trainable network for Chinese LP detection and recognition, which almost reaches a trade-off between accuracy and efficiency as a baseline. Through utilizing the common feature extraction branch and the RRoIAlign~\cite{fots} operator, end-to-end training is achieved, and the error accumulation problems are alleviated with a real-time efficiency kept.
		\item Our code and dataset will be publicly available soon. To facilitate the reference of researchers and get more progress, we will upload related materials.
	\end{itemize}
	
	\section{Related Work}
	\label{sec:related}
	\subsection{LP Datasets}
	Due to the importance of LP detection and recognition, researchers built and published a number of LP datasets. ReId~\cite{reid} is a dataset for license plate recognition with 76k images gathered from surveillance cameras on highway toll gates. Caltech~\cite{caltech} and Zemris~\cite{zemris} collected over 600 images from the road and freeways with high-resolution cameras. Hsu~\textsl{et al.}~\cite{aolp} presented a dataset for applications of access control, traffic law enforcement and road patrol. Gon{\c{c}}alves~\textsl{et al.}~\cite{ssig} proposed Sense SegPlate Database to evaluate license plate character segmentation problem. Laroca~\textsl{et al.}~\cite{ufpr} provided a dataset that includes 4,500 fully annotated images from 150 vehicles in real-world scenarios. These datasets strongly support the researches of LP detection and recognition methods.
	
	However, LPs in different countries and regions are usually not the same. The mentioned datasets mostly contain LPs that only include alphanumeric characters. In some countries or regions, such as Chinese mainland regions, Japan and South Korea, the LPs contain some special characters. Therefore, it is still significant to build new datasets for these LPs. Among them, mainland Chinese LP detection and recognition are one of the important tasks which require a large amount of data. To meet the demand, Chinese LP datasets were proposed. Zhou~\textsl{et al.}~\cite{pvw} collected 220 LP images where the LPs were of little affine distortion. Yuan~\textsl{et al.}~\cite{pku} presented a dataset that contains vehicle images captured from various scenes under diverse conditions. Zhang~\textsl{et al.}~\cite{clpd} proposed a dataset that contains 1,200 LP images from all 31 provinces in mainland China. CCPD~\cite{ccpd} is a large and comprehensive dataset with about 290k images that contain plates with various angles, distances, and illuminations. Though they made important contributions to the progress of detection and recognition methods, there are still not enough large and representative datasets.
	
	\subsection{LP Detection and Recognition}
	Owing to the successes of text detection and recognition, LP processing is also well developed.
	There are methods with end-to-end frameworks, which achieved excellent performance. Zhang~\textsl{et al.}~\cite{vlpdr} integrated LP detection, tracking, and recognition into a unified framework via deep learning. Silva~\textsl{et al.}~\cite{cnnlpdr} proposed to identify the vehicle and the LP region using two passes on the same CNN and then to recognize the characters using a second CNN. Kessentini~\textsl{et al.}~\cite{twostage} presented a two-stage network to achieve the detection and recognition of LPs. In~\cite{mtlpr}, a light CNN was proposed for detection and recognition, which achieved real-time efficiency.
	
	As for Chinese LPs, there are also excellent end-to-end frameworks. Laroca~\textsl{et al.}~\cite{yoloalpr} proposed a unified approach for LP detection and layout classification to improve the recognition results.  Qin~\textsl{et al.}~\cite{eulpr} proposed a unified method that can recognize both single-line and double-line LPs in an end-to-end way without line segmentation and character segmentation. However, Chinese LP processing is still challenging due to the large number of categories of Chinese characters. Meanwhile, LP processing under unconstrained scenarios also faces many problems. Therefore, it is still valuable to propose a new end-to-end framework for Chinese LP detection and recognition.
	
	\section{CRPD Overview}
	\label{sec:1}
	\subsection{Constitution of Data}
	\label{sec:2}
	
	\begin{figure}[h]
		\centering
		\includegraphics[width=2.7in]{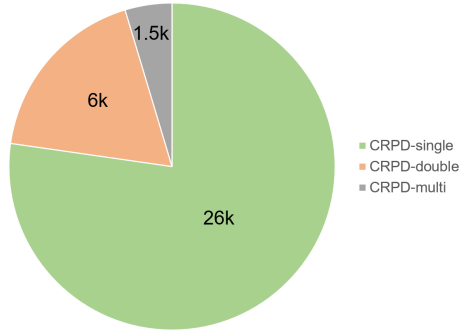}
		\caption{The constitution of CRPD dataset.}
		\label{cons}
	\end{figure}
	
	Because CRPD is presented as a supplement for existing datasets, special attention is paid to the diversity of data. The images are mainly captured on electronic monitoring systems, including vehicles that are running, turning, parked, or far away which may cause blur and rotation. The scene includes day and night and different weathers. Quite a part of the images contains more than one LP from different provinces and cities, and there are LPs of special vehicles involved, such as coach cars, police cars, and trailers, whose LPs will contain some special characters. The dataset includes three sub-datasets according to the numbers of LPs: CRPD-single, CRPD-double, and CRPD-multi, as shown in Figure~\ref{cons}. CRPD-single contains images with only one LP, CRPD-double contains images with two LPs, and CRPD-multi contains images with three or more LPs.
	
	There are totally about 25k images for training, 6.25k for validating, and 2.3k for testing. In CRPD-single, there are 20k images for training, 5k for validating, and 1k for testing. In CRPD-double, there are 4k images for training, 1k for validating, and 1k for testing. In CRPD-multi, there are 1k images for training, 0.25k for validating, and 0.3k for testing.
	
	The annotations consist of three parts. The first is LP content which includes numbers, Chinese and English characters. There are some LPs that are too small or seriously blurred whose content is unidentifiable, and they are also annotated while the unrecognizable characters are replaced with a special one. The second is the coordinate of four vertices of the LPs. The last is the LP type, including blue (small cars), yellow and single line (front of large cars), yellow and double lines (back of large cars), and white (police cars).
	
	\subsection{Data Analysis}
\begin{figure}[h]
		\centering
		\includegraphics[width=3.5in]{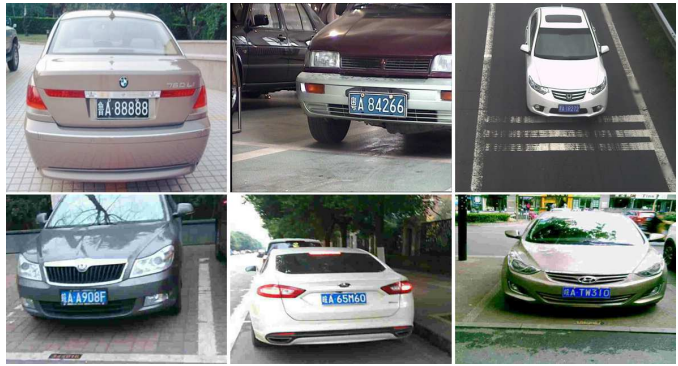}%
		\caption{Examples of LP images in existing public datasets.}
		\label{public_datasets}
	\end{figure}

\begin{figure}[t]
		\centering
		\includegraphics[width=3.5in]{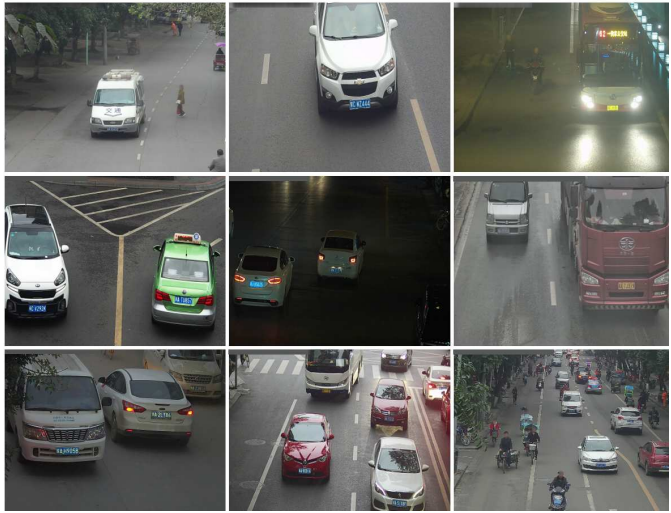}%
		\caption{Examples of LP images in CRPD-single (the first row), CRPD-double (the second row), and CRPD-multi (the third row).}
		\label{crpd}
	\end{figure}

	CRPD provides more than 30k LP images with annotations. Though it is not the largest dataset in current frequently used Chinese datasets, some characteristics of CRPD will be helpful for training a robust model. Existing datasets mostly contain single and focused LP in one image, as shown in Figure~\ref{public_datasets}. In some cases, such as electronic toll payment or parking managing systems, the data is highly effective. But when dealing with data for traffic monitoring systems, suspect car tracking, or vehicle flow measuring, images with more LPs will be required. CRPD aims to fill up this deficiency, so we paid special attention to the number of LPs, as shown in Figure~\ref{crpd}.
	
	Also, CRPD has some other advantages for building a robust model. To illustrate them, we compare with CCPD~\cite{ccpd}, EasyPR~\cite{easypr}, ChineseLP~\cite{pvw} and CLPD~\cite{clpd} in some aspects, which are shown in Tables~\ref{LP_number},~\ref{status} and~\ref{type}.
	
	The first is the number of LPs. Images in CCPD~\cite{ccpd} and CLPD~\cite{clpd} contain one LP, which can better indicate the detection Precision of a network. But as noted above, this may restrict the scenarios where it can be used. In comparison, EasyPR~\cite{easypr}, ChineseLP~\cite{pvw} and our CRPD have better compatibility in this aspect.
	
	The second is the status of vehicles. CCPD~\cite{ccpd} contains images captured from parking lots, and they are more interested in LPs in various circumstances with different illumination, rotation, and blur. Therefore, the vehicle status is not focused. Our CRPD concentrates on the capability to deal with a variety of vehicles, so there is better coverage on vehicle status.
	
	\begin{table}[!t]
		\caption{A comparison on the number of images with different numbers of LPs between current public Chinese LP datasets.}
		\label{LP_number}
		\centering
		\begin{tabular}{cccccc}
			\toprule[0.75pt]
			LP Number  &  CCPD  & EasyPR & ChineseLP & CLPD & CRPD    \\
			\midrule[0.5pt]	
			$=1$       & 224001 &   225  & 392& 1200&  26659  \\
			$=2$       &   0    &   21   &  16&    0&  6242   \\
			$=3$       &   0    &   10   &   1&    0&  1232   \\
			$\geq4$    &   0    &   0    &   2&    0&  371    \\
			\bottomrule[0.75pt]
		\end{tabular}
	\end{table}
	
	\begin{table}[t]
		\caption{A comparison of the coverage of different vehicle statuses between current public Chinese LP datasets.}
		\label{status}
		\centering
		\begin{tabular}{cccccc}
			\toprule[0.75pt]
			Status of Vehicles & CCPD      & EasyPR   & ChineseLP& CLPD& CRPD      \\
			\midrule[0.5pt]
			Parked             & \ding{52} &\ding{52} &\ding{52} &\ding{52} &\ding{52}  \\
			Running            & \ding{55} &\ding{52} &\ding{52} &\ding{52} &\ding{52}  \\
			Turning            & \ding{55} &\ding{52} &\ding{52} &\ding{52} &\ding{52}  \\
			Far away           & \ding{55} &\ding{55} &\ding{55} &\ding{52} &\ding{52}  \\
			\bottomrule[0.75pt]
		\end{tabular}
	\end{table}
	
\begin{figure}[!t]
		\centering
		\includegraphics[width=3.5in]{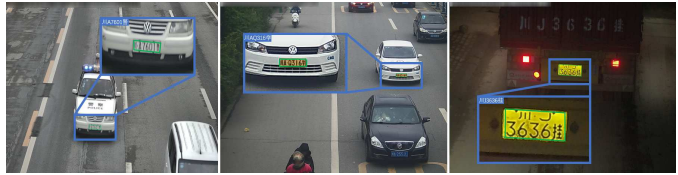}%
		\caption{LPs of special vehicles with their annotations in CRPD. The green rectangles are the annotated boxes and the text on the top left corner of the zooming rectangle is the annotated LP content.}
		\label{special}
	\end{figure}

	The last is the vehicle types. Though there are not a number of special vehicles on the road, the detection and recognition of them are still of great importance. Thus, we paid attention to the LPs of special vehicles to ensure the capability of the network trained by CRPD to deal with these LPs. Some images of them are shown in Figure~\ref{special}. Altogether, our CRPD covers most a variety of common scenes. It is a strong supplement to Chinese LP datasets either used as training or evaluating data.
	
	\begin{table}[t]
		\caption{A comparison of the coverage of different vehicle types between current public Chinese LP datasets.}
		\label{type}
		\centering
		\begin{tabular}{cccccc}
			\toprule[0.75pt]
			Type of Vehicles    & CCPD      & EasyPR    & ChineseLP& CLPD& CRPD         \\
			\midrule[0.5pt]
			Coach Vehicles      & \ding{55} & \ding{52} & \ding{55}& \ding{52}&  \ding{52}   \\
			Police Vehicles     & \ding{55} & \ding{55} & \ding{55}& \ding{52}&  \ding{52}   \\
			% Military Vehicles  & \ding{55} & \ding{55} &  \ding{52}   \\
			Trailers            & \ding{55} & \ding{55} & \ding{55}& \ding{55}&  \ding{52}   \\
			\bottomrule[0.75pt]
		\end{tabular}
	\end{table}
	
	However, there are also some limitations of CRPD. First, the location of each LP character is not annotated, which restricts the applications of object detection and data augmentation. Meanwhile, it is also difficult to detect each character because it is very small in the perspective of electronic monitoring systems. Second, the LPs are all on the vehicles. The actual LPs can be held in hand or placed on the ground in some circumstances, and the detection and recognition of these LPs are also useful.
	
	\begin{figure*}[t]
		\centering
		\includegraphics[width=5in]{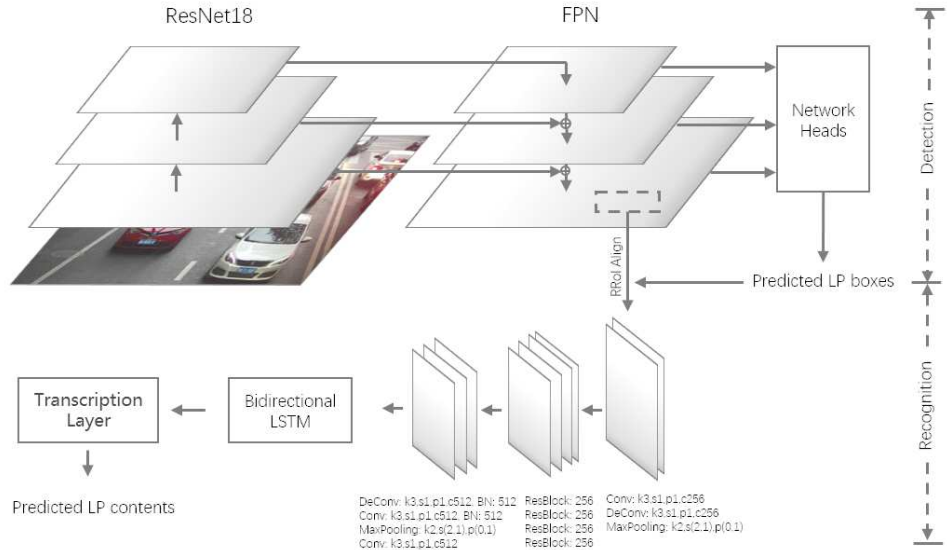}%
		\caption{The framework of our network. Conv, DeConv, Bn, MaxPooling, and ResBlock stand for convolution layer, deformable convolution layer, batch normalization layer, max pooling layer, and residual block, respectively. k, s, p, and c stand for kernel size, stride, padding size, and output channel number, respectively, with the size behind each of them.}
		\label{pipeline}
	\end{figure*}
	
	\section{Methodology}
	
	\label{methodology}
	
	In this section, we will describe the proposed method in detail, and the pipeline is shown in Figure~\ref{pipeline}.
	
	\subsection{Detection Branch}
	\label{methodology_det}
	
	The first stage is the LP detection step. As noted above, there are some obstacles for LP detection networks to reach a balance between accuracy and efficiency. Considering the trade-off, we utilize our previous STELA~\cite{stela}, a totally real-time detector, as the basis of our detection branch. The detection network is implemented on RetinaNet~\cite{retinanet} and utilizes Feature Pyramid Network (FPN)~\cite{fpn} to construct a rich, multi-scale feature pyramid from a single resolution input image. It consists of three portions: anchor classification, rotated bounding box regression, and anchor refining.
	
	\subsubsection{Anchor Classification}
	\label{methodology_det_cls}
	
	As we do not generate region proposals, class imbalance problems still exist in our scheme. That means there are only a few anchors that are annotated as positive (the object), while the others are negative. Therefore, Focal Loss~\cite{retinanet} are utilized to calculate the loss of classification, as it is designed to deal with the problem. Firstly, we define $p_t$ with
	\begin{equation}
		p_t=
		\begin{cases}
			p& if~y=1\\
			1-p& otherwise
		\end{cases}
	\end{equation}
	where $p\in [0,1]$ is the predicted probability and $y=1$ specifies the ground-truth class. Then the loss is defined as 
	\begin{equation}
		L_{cls} = FL(p_t)=-\alpha_{t}(1-p_{t})^{\gamma}log(p_t)
	\end{equation}
	$\alpha_{t}$ is a balanced weighting factor and $\gamma$ is a focusing parameter. They are set to 0.25 and 2.0 respectively, which is the same as the original Focal Loss~\cite{retinanet}.
	
	\begin{figure}[b]
		\centering
		\includegraphics[width=2.3in]{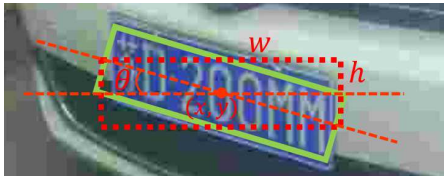}%
		\caption{Illustration of the parameters. The red dotted box represents $b$ which is a box and the green solid box represents $g$ which is the groundtruth.}
		\label{regression}
	\end{figure}
	
	\subsubsection{Rotated Bounding Box Regression}
	\label{methodology_det_reg}
	
	For the detection of tilted LP, we utilize rotated bounding boxes to match the instances. The box can be represented by a five tuple $(x,y,w,h,\theta)$, in which $x$ and $y$ are the coordinate of the center point, $w$ and $h$ are the width and height of the box, and $\theta$ is the angle to horizontal, as shown in Figure~\ref{regression}. For the regression operation, the distance vector $\Delta=(\delta_x,\delta_y,\delta_w,\delta_h,\delta_{\theta})$ is defined as
	\begin{equation}
		\delta_x = (g_x -b_x )/b_w ,  ~\delta_y = (g_y -b_y )/b_h
	\end{equation}
	\begin{equation}
		\delta_w = log(g_w /b_w ),  ~\delta_h = log(g_h / b_h)
	\end{equation}
	\begin{equation}
		\delta_{\theta}=tan(g_{\theta})-tan(b_{\theta})
	\end{equation}
	where $b$ and $g$ represent a bounding box and the corresponding target groundtruth respectively. The loss of the regression can be calculated by
	\begin{equation}
		L_{loc} = smooth_{L_1}(\Delta_{t}-\Delta_{p})
	\end{equation}
	where $smooth_{L_1}$ is the smooth L1 loss~\cite{fastrcnn}, $\Delta_t$ is the target and $\Delta_p$ is the predicted tuple.
	
	\subsubsection{Learned Anchor}
	\label{methodology_det_ref}
	
\begin{figure}[t]
		\centering
		\includegraphics[width=3.5in]{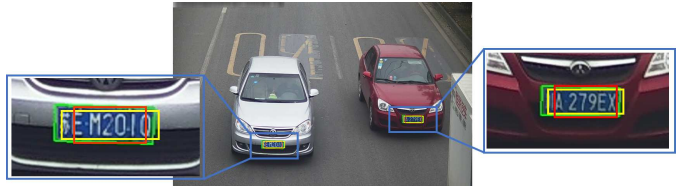}%
		\caption{The illustration of anchors and boxes. The red, yellow, and green boxes represent the original anchor, the learned anchor, and the final output boxes respectively. The original anchor (red) and learned anchor (yellow) have the same center point.}
		\label{anchors}
	\end{figure}

	As depicted in~\cite{stela}, the most important part of the proposed scheme in two-stage is that the selected proposals are chosen by learning. The manually-defined original anchors with fixed scales and aspect ratios may not be the optimal designs, so an extra regression branch for anchor refining is added. The final classification and regression task will be reached on the learned anchors, which brings an improvement in accuracy with a little increment of computation. The original anchor, learned anchor, and output boxes are illustrated in Figure~\ref{anchors}. It is obvious that the center should be well aligned with the pixel in feature maps. Thus, the offsets are only regressed within $\Delta'=(\delta'_{w},\delta'_{h},\delta'_{\theta})$, which means that only the shapes are adjusted. The loss can be calculated with
	\begin{equation}
		L_{ref} = smooth_{L_1}(\Delta'_{t}-\Delta'_{p})
	\end{equation}
	in which $\Delta'_t$ is the target and $\Delta'_p$ is the predicted tuple. And finally, the total loss is
	\begin{equation}
		L_{det} = \lambda_{ref}L_{ref}+\lambda_{loc}L_{loc}+\lambda_{cls}L_{cls}
	\end{equation}
	in which $\lambda_{ref}, \lambda_{loc}$ and $\lambda_{cls}$ are the weights which are set to 0.5, 0.5, and 1 which have been proven to be effective.
	
	\subsection{Recognition Branch}
	\label{methodology_rec}
	
	The recognition step is the second stage of the LP processing tasks. Because most of the networks for recognition achieved high efficiency, we simply utilize modules based on CRNN~\cite{crnn} to achieve the recognition. There are three parts in the modules: the convolutional layers, the recurrent layers, and the transcription layer.
	
	In consideration that the backbone of our network has already extracted critical features of the input images, to avoid redundant computation, we utilize the processed feature maps as the input of the recognition branch. In order to effectively deal with rotated plates, RRoIAlign~\cite{fots} that can crop the feature map with a rotated box, is applied to crop the maps according to the groundtruth boxes of LPs, and the cropped size is $8\times 25$. While training, we consider that the feature maps may not be processed well when the predicted boxes are not accurate, so the feature maps will also be cropped with both groundtruth boxes and predicted boxes with a score higher than 0.9.
	
\begin{figure}[h]
		\centering
		\includegraphics[width=2.7in]{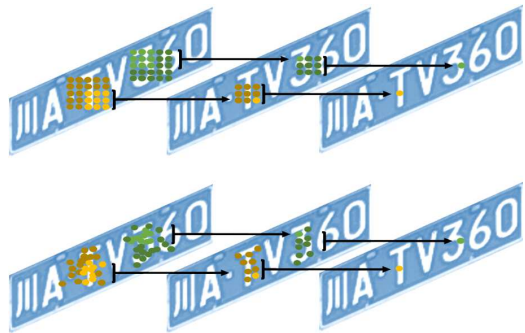}
		\caption{Indication of fixed receptive fields in standard convolution (the first row) and adaptive receptive fields in deformable convolution (the second row). In each image triplet, the left shows the sampling locations of two levels of $3\times3$ filters on the preceding feature map, the middle shows the sampling locations of a $3\times3$ filter, and the right shows two activation units. Two sets of locations are highlighted according to the activation units.}
		\label{deformable}
	\end{figure}

	And because the input is feature maps, we remove the first three convolutions of the original convolutional layers of CRNN~\cite{crnn} to avoid the over-fitting problems. And in order to reach better accuracy, we refer to our another previous work~\cite{drnn}, and replace two convolution layers with deformable convolution layers and add four residual blocks in this branch. The deformable convolution layers have an adaptive receptive field that can better cover the text area, as shown in Figure~\ref{deformable}.
	
	The architecture of the convolutional layers is shown in Figure~\ref{pipeline}. The recurrent layers are based on bidirectional LSTM~\cite{lstm}, and the transcription component is based on CTCLoss~\cite{ctc}. The total loss function is a weighted loss
	
	\begin{equation}
		L=\lambda_{det} L_{det} +\lambda_{rec} L_{rec}
	\end{equation}
	
	where $\lambda_{det}$ and $\lambda_{rec}$ are constants that indicate the strength of the detection and recognition modules. And in our training, the value of $L_{rec}$ is one-magnitude-order larger than $L_{det}$, so to keep a balance, we set them to 1 and 0.1, respectively.

	\section{Experiments}
	\label{experiments}
	\subsection{Training Details}
	\label{experiments_details}
	
	The training and testing datasets are from CCPD~\cite{ccpd}, EasyPR~\cite{easypr}, and our CRPD. The input images are resized to $640\times 640$ with three channels. In the training stage, the optimizer of the network is Adam [18], the batch size is set to 32, and the learning rate is 1e-4. The network is trained for 35000 iterations which consume about 10 hours. The proposed method is implemented by PyTorch~\cite{pytorch}. The experiments are carried on a platform with Intel Xeon(R) E5-2630 v3 CPU and a single NVIDIA TITAN RTX GPU.
	
	\subsection{Evaluation Metrics}
	\label{experiments_metrics}
	
	To demonstrate the effectiveness of the methods, we utilize the protocols described in~\cite{ccpd} to evaluate the models. The detection will be considered as a match if it overlaps a ground truth bounding box by more than 60\% and the words match exactly. Then Recall, Precision, and F-score are calculated with
	\begin{equation}
		Recall=\frac{TP}{TP+FN}
	\end{equation}
	\begin{equation}
		Precision=\frac{TP}{TP+FP}
	\end{equation}
	\begin{equation}
		F-score=\frac{2\times Precision \times Recall}{Precision+Recall}
	\end{equation}
	where $TP$ represents the number of positive objects that are predicted as positive and the words match, $FP$ represents the number of negative objects that are predicted as positive, and $FN$ represents the number of negative objects that are predicted as negative.  
	
	\subsection{Ablations}
	\label{experiments_ablations}
	
	To demonstrate that the unified architecture brings some improvements and our scheme reaches the best performance, we evaluate the proposed components. Firstly, we test the effectiveness of our unified network. As comparisons, cascaded STELA~\cite{stela} and CRNN~\cite{crnn} models are utilized. The two networks are trained respectively, and the CRNN~\cite{crnn} model will process the original images cropped in the light of the boxes predicted by STELA~\cite{stela}. All the models are trained on CRPD and evaluated on CCPD~\cite{ccpd}, EasyPR~\cite{easypr}, and CRPD. From Table~\ref{unified_expr}, we see that our network achieves a few improvements. A better result is reached on CRPD because the training and testing data have the same distribution. The LPs in CCPD~\cite{ccpd} have a larger size, so the model trained on our CRPD, which contains LPs with small sizes, is not able to reach the best performance.
	
	\begin{table}[h]
		\centering
		\caption{Ablations of the unified scheme on different datasets.}
		\label{unified_expr}
		\begin{threeparttable}
		\begin{tabular}{p{1.3cm}<\centering p{0.55cm}<\centering p{0.55cm}<\centering p{0.55cm}<\centering c p{0.55cm}<\centering p{0.55cm}<\centering p{0.55cm}<\centering}
			\toprule[0.75pt]
			\multirow{2}{*}{Dataset}   & \multicolumn{3}{c}{STELA+CRNN}   & & \multicolumn{3}{c}{Our Method} \\
			\cmidrule[0.5pt]{2-4}\cmidrule[0.5pt]{6-8}
			&  R  & P & F & & R  & P & F \\
			\midrule[0.5pt]
			CCPD     & 79.1          &   67.8          & 73.0   &       &  75.6          &  72.1          & 73.8            \\
			EasyPR & \textbf{90.2} &   72.8          & 80.6        &  &  89.9          &  73.0          & 80.6            \\
			CRPD                 & 88.3          &   \textbf{82.9} & \textbf{85.5} & & \textbf{95.4} &  \textbf{84.1} & \textbf{89.4}   \\
			\bottomrule[0.75pt]
		\end{tabular}
		\begin{tablenotes}
			\item R: Recall; P: Precision; F: F-score
		\end{tablenotes}
		\end{threeparttable}
	\end{table}
	
	Then as we utilize the predicted boxes of the detection modules to crop the feature maps in the training stage, we evaluate the effectiveness of them with different scores. In Table~\ref{ablations_predbox}, it is obvious that utilizing predicted boxes will bring an improvement on the Recall because more incomplete LPs will be detected, with a little descend on Precision. Considering the recognition branch will also become more robust with this mechanism, it will make the network able to better deal with inaccurate boxes. But because Chinese LPs usually have seven characters, a box with a score lower than 0.86 may cut a whole character off. Thus, there must be a deterioration of the performance when using boxes with scores only higher than 0.85. And the number of boxes with scores higher than 0.95 is less, which causes a smaller batch size for recognition modules, so we choose 0.9 as the threshold finally.
	
	\begin{table}[h]
		\caption{Ablations on CRPD-all between different thresholds of scores of the predicted boxes utilized to train the model.}
		\label{ablations_predbox}
		\centering
		\begin{tabular}{cccc}
			\toprule[0.75pt]
			Threshold            &  Recall   & Precision &  F-score        \\
			\midrule[0.5pt] 
			None                 &  91.2          &  \textbf{85.4} &   88.2         \\
			\textgreater0.95     &  91.7          &  84.6          &   88.0         \\
			\textgreater0.90     &  \textbf{95.4} &  84.1          & \textbf{89.4}  \\
			\textgreater0.85     &  94.5          &  82.5          &   88.1         \\
			\bottomrule[0.75pt]
		\end{tabular}
	\end{table}
	
	The way to crop the feature maps may also bring different results of the recognition modules. We compare RoIPool~\cite{fastrcnn}, RoIAlign~\cite{maskrcnn}, and RRoIAlign~\cite{fots}, and the results are shown in Table~\ref{ablations_crop}. There is no obvious difference between the first two methods, and we think the reason is that LPs are not small objects, so some slight errors will not influence much. RRoIAlign~\cite{fots} can better handle the work because the recognition branch will have great progress with consideration of the rotation degrees.
	
	\begin{table}[t]
		\caption{Ablations on CRPD-all between different region feature extracting approaches.}
		\label{ablations_crop}
		\centering
		\begin{tabular}{cccc}
			\toprule[0.75pt]
			Approach    &  Recall   & Precision &  F-score       \\
			\midrule[0.5pt]   
			RoIPooling    &  95.0          & 80.2          &  87.0           \\
			RoIAlign      &  \textbf{95.7} & 80.1          &  87.2           \\
			RRoIAlign     &  95.4          & \textbf{84.1} & \textbf{89.4}   \\
			\bottomrule[0.75pt]
		\end{tabular}
	\end{table}
	
	In the recognition branch, we crop the feature maps yield by different layers of the FPN~\cite{fpn} components to explain why we utilize those from the third layer. Though it seems to utilize the feature maps from the third layer is optimal in Table~\ref{ablations_fpnlayer}, but we consider it is because most of the LPs in the images of CRPD have sizes which match the anchors in the third layer. In the fifth or deeper layers, the anchors are mappings of some huge boxes in the original images, but most of the LPs have a small size.
	
	\begin{table}[h]
		\caption{Ablations on CRPD-all with feature maps from different FPN layers to train the model.}
		\label{ablations_fpnlayer}
		\centering
		\begin{tabular}{cccc}
			\toprule[0.75pt]
			Feature Map Layer  &  Recall  & Precision & F-score       \\
			\midrule[0.5pt] 
			P3            & \textbf{95.4} & \textbf{84.1} &  \textbf{89.4}  \\
			P4            & 82.4          & 70.1          &  75.8           \\
			P5            & 2.6           & 2.9           &  2.7            \\
			\bottomrule[0.75pt]
		\end{tabular}
	\end{table}
	
	\begin{table}[h]
		\caption{Ablations on CRPD-all with feature maps from different FPN layers to train the model.}
		\label{ablations_deformable}
		\centering
		\begin{tabular}{cccc}
			\toprule[0.75pt]
			Deformable Conv  &  Recall  & Precision & F-score         \\
			\midrule[0.5pt]
			\ding{55}        & 93.0           & 83.7          &  88.1          \\
			\ding{52}        & \textbf{95.4}  & \textbf{84.1} & \textbf{89.4}  \\
			\bottomrule[0.75pt]
		\end{tabular}
	\end{table}
	
	Finally, to demonstrate that the deformable layers and residual blocks in the recognition branch are effective, ablations are involved, as shown in Table~\ref{ablations_deformable}. In our metrics, only when the content matches exactly, the result will be regarded as correct. Therefore, deformable convolution brings an improvement in recognition, and the Recall and Precision are both improved.
	
	\subsection{Comparisons}
	\label{experiments_comp}
	
	\begin{table*}[h]
		\scriptsize
		\caption{Comparisons on CRPD between our method and other methods. The methods that are based on deep-learning are trained on CRPD.}
		\label{table_crpd_comp}
		\centering
		\begin{threeparttable}
		\begin{tabular}{m{2.9cm}<{\centering}m{0.5cm}<{\centering}m{0.8cm}<{\centering}m{0.9cm}<{\centering}m{0.3cm}<{\centering}cm{0.5cm}<{\centering}m{0.8cm}<{\centering}m{0.9cm}<{\centering}m{0.3cm}<{\centering}}
			\toprule[0.75pt]
			\multirow{2}{*}{Method}& \multicolumn{4}{c}{CRPD-all}    &                                & \multicolumn{4}{c}{CRPD-single} \\
			\cmidrule[0.5pt]{2-5}\cmidrule[0.5pt]{7-10}
			& Recall& Precision& F-score& FPS& &  Recall& Precision& F-score& FPS\\
			\midrule[0.5pt]
			EasyPR&    2.0&  1.3&  1.6&  6& &  2.3&  1.3&  1.7&  6\\
			SSD512 + CRNN& \textbf{97.8}& 27.2& 42.6& \textbf{66}& & \textbf{98.9}& 28.7& 44.4& \textbf{71}\\
			     YOLOv3 + CRNN& 73.0& 61.0& 66.5& 18& & 73.7& 59.4& 65.8& 18\\
			     YOLOv4 + CRNN& 84.4& 60.5& 70.5& 40& & 87.3& 68.4& 76.7& 40\\
			    SYOLOv4 + CRNN& 86.8& 71.0& 78.2& 35& & 90.1& 72.4& 80.3& 35\\
			Faster-RCNN + CRNN& 79.9& 73.7& 76.7& 19& & 81.4& 71.7& 76.3& 20\\
			      STELA + CRNN& 88.3& 82.9& 85.5& 36& & 83.1& 73.3& 77.9& 36\\
			Ours& 95.4& \textbf{84.1}& \textbf{89.4}& 30& & 96.3& \textbf{83.6}& \textbf{89.5}& 35\\
			\hline
			\hline
			\multirow{2}{*}{Method}  & \multicolumn{4}{c}{CRPD-double} & & \multicolumn{4}{c}{CRPD-multi} \\
			\cmidrule[0.5pt]{2-5}\cmidrule[0.5pt]{7-10}
			&  Recall& Precision& F-score& FPS& & Recall& Precision& F-score& FPS\\
			\hline
			EasyPR& 1.5& 1.2& 1.3& 6& & 1.8& 1.7& 1.8& 6\\
			SSD512 + CRNN& \textbf{97.5}& 27.5& 42.9& \textbf{66}& & \textbf{93.5}& 21.1& 34.5& \textbf{63}\\
			     YOLOv3 + CRNN& 74.6& 64.4& 69.1& 17& & 66.2& 61.3& 63.6& 17\\
			     YOLOv4 + CRNN& 90.4& 41.2& 56.6& 40& & 88.9& 36.8& 52.0& 39\\
			    SYOLOv4 + CRNN& 91.5& 75.9& 83.0& 35& & 91.5& 75.2& 82.5& 35\\
			Faster-RCNN + CRNN& 81.1& 77.9& 79.5& 19& & 69.3& 75.2& 72.1& 17\\
			      STELA + CRNN& 84.0& 80.6& 82.3& 34& & 77.6& 82.8& 80.1& 33\\
			Ours& 95.8& \textbf{84.5}& \textbf{89.8}& 30& & 90.8& \textbf{85.0}& \textbf{87.7}& 26\\
			\bottomrule[0.75pt]
		\end{tabular}
		\begin{tablenotes}
			\item SYOLOv4: Scaled YOLOv4
		\end{tablenotes}
		\end{threeparttable}
	\end{table*}
	
	To demonstrate that our method has reached a satisfactory performance, we make some comparisons with other methods. We involve EasyPR~\cite{easypr}, SSD~\cite{ssd}, YOLO-v3~\cite{yolov3}, YOLO-v4~\cite{yolov4}, Scaled YOLO-v4~\cite{scaledyolov4} and Faster-RCNN~\cite{fasterrcnn}. A CRNN~\cite{crnn} model is appended to achieve recognition. We utilize cascaded STELA~\cite{stela} and CRNN~\cite{crnn} as our baseline in experiments. From Table~\ref{table_crpd_comp}, it can be observed that EasyPR~\cite{easypr} cannot treat LPs in CRPD well because it depends on manually designed features. SSD~\cite{ssd}, YOLOv3~\cite{yolov3}, YOLO-v4~\cite{yolov4}, Scaled YOLO-v4~\cite{scaledyolov4} and Faster-RCNN~\cite{fasterrcnn} are not specially designed for LPs or text, but the results are still competitive. Our method reaches the best on all the sub-datasets, which proves the effectiveness.
	
	Because Xu~\textsl{et al.}~\cite{ccpd} evaluated their method on different circumstances, for fair comparisons, experiments of our method on these datasets are also utilized. And as the other methods are trained on CCPD~\cite{ccpd}, we also utilize it as the training data of our method in this comparison. Because images in CCPD~\cite{ccpd} only contain one LP in one image, only Precision is involved when the Recall is not considered. Following the experiments of Xu~\textsl{et al.}~\cite{ccpd}, Cascade classifier~\cite{wang2007cascade}, SSD300~\cite{ssd}, YOLO9000~\cite{yolov2}, Faster-RCNN~\cite{fasterrcnn} are involved as the detector with Holistic-CNN~\cite{reid} as the recognition model. And end-to-end methods TE2E~\cite{te2e} and RPNet~\cite{ccpd} are utilized for comparisons. We also involve YOLO-v4~\cite{yolov4} and Scaled YOLO-v4~\cite{scaledyolov4} with CRNN~\cite{crnn} to report more results. Our baseline which consists of STELA~\cite{stela} and CRNN~\cite{crnn} are also involved. From Table~\ref{table_ccpd_comp}, it can be observed that our method has reached the best in most circumstances, and the efficiency is also competitive.
	
	\begin{table*}[h]
		\caption{Comparisons of Precision on CCPD between our method and others.}
		\scriptsize
		\label{table_ccpd_comp}
		\centering
		\begin{threeparttable}
		\begin{tabular}{m{2.95cm}<{\centering}m{0.3cm}<{\centering}m{0.3cm}<{\centering}m{0.4cm}<{\centering}m{0.3cm}<{\centering}m{0.4cm}<{\centering}m{0.6cm}<{\centering}m{0.4cm}<{\centering}m{0.8cm}<{\centering}m{0.9cm}<{\centering}m{0.3cm}<{\centering}} 
			\toprule[0.75pt]
			Method                                                            
			& Size&  AP& Base&   DB&   FN& Rotate& Tilt& Weather& Challenge& FPS\\
			\midrule[0.5pt]
			Cascade classifier + HC
			& 480& 58.9& 69.7& 67.2& 69.7&    0.1&  3.1&    52.3&      30.9&  29\\
			SSD300 + HC
			& 300& 95.2& 98.3& 96.6& 95.9&   88.4& 91.5&    87.3&      83.8&  35\\
			YOLO9000 + HC
			& 480& 93.7& 98.1& 96.0& 88.2&   84.5& 88.5&    87.0&      80.5&  36\\
			YOLOv4 + CRNN
			& 512& 94.7& 97.8& 94.6& 87.3&   82.9& 89.9&    83.3&      75.7&  40\\
			SYOLOv4 + CRNN
			& 640& 95.3& 97.8& 95.0& 88.9&   84.9& 91.5&    90.4&      77.1&  34\\
			Faster-RCNN + HC
			& 600& 92.8& 97.2& 94.4& 90.9&   82.9& 87.3&    85.5&      76.3&  13\\
			TE2E
			& 600& 94.4& 97.8& 94.8& 94.5&   87.9& 92.1&    86.8&      81.2&   3\\
			RPNet
			& 480& 95.5& \textbf{98.5}& 96.9& 94.3& 90.8& 92.5& 87.9& 85.1&  \textbf{61}\\
			STELA + CRNN
			& 640& 97.8& 97.9& \textbf{98.3}& 94.5& 90.1& 91.3& 89.5& 83.6&  41\\
			Ours
			& 640& \textbf{97.9}& 98.3& 98.0& \textbf{97.2}& \textbf{92.5}& \textbf{93.7}& \textbf{90.7}& \textbf{87.9}& 30\\
			\bottomrule[0.75pt]
		\end{tabular}
		\begin{tablenotes}
			\item SYOLOv4: Scaled YOLOv4; HC: Holistic-CNN; AP: average percent of all the circumstances
		\end{tablenotes}
		\end{threeparttable}
	\end{table*}
	
\begin{figure*}[t]
		\centering
		\includegraphics[width=4.5in]{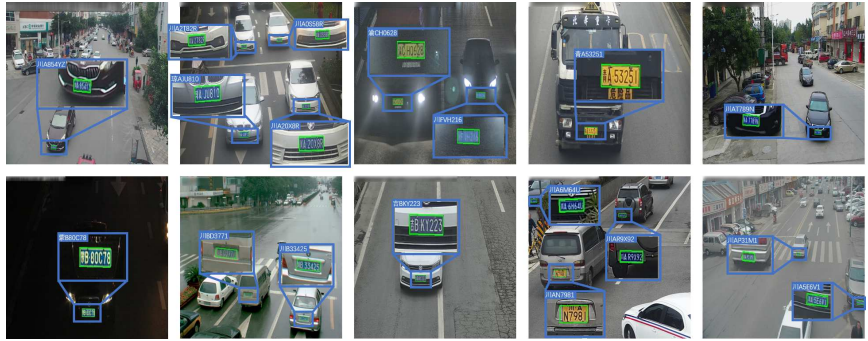}%
		\caption{Spotting results of our method on CRPD. The green rectangles are the predicted rotated bounding boxes and the text on the top left corner of the zooming rectangle is the predicted text.}
		\label{results}
	\end{figure*}

\begin{figure*}[h]
		\centering
		\includegraphics[width=4.5in]{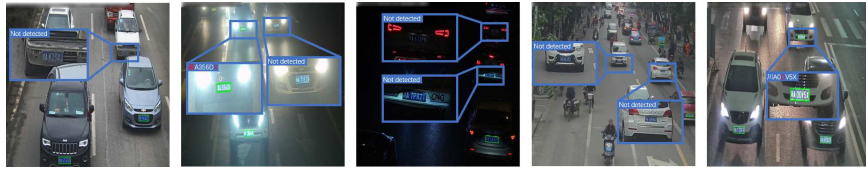}%
		\caption{Examples of failed recognition on CRPD. The green rectangles are the predicted rotated bounding boxes and the text on the top left corner of the zooming rectangle is the description of the LP, and the red characters are the ones that are recognized incorrectly. For simplicity, only the failed examples are zoomed in.}
		\label{results_wrong}
	\end{figure*}

	Finally, some detection and recognition results on CRPD are shown in Figure~\ref{results} and~\ref{results_wrong}. In the illustrations, we see that most LPs that are recognized incorrectly are also hard to distinguish by humans. Our proposed method is still effective in most cases, and the performance is competitive.
	
	\section{Discussion and Conclusion}
	\label{conclusion}
	
	In this paper, we present a dataset with Chinese LP images, which is named CRPD. As a supplement to multi-LP datasets, CRPD includes three sub-datasets, CRPD-single, CRPD-double, and CRPD-multi, which are able to deal with a variety of application scenarios. And CRPD covers many kinds of vehicles and a number of environments that will be helpful to build a robust model. We also propose an end-to-end trainable network to detect and recognize LPs with high efficiency as the baseline of the dataset. The experiments demonstrate the effectiveness of our proposed components, and the performance of the network is satisfactory. In the future, we hope CRPD will become a new benchmark on multi-LP detection and recognition tasks. We also consider utilizing the network for end-to-end scene text spotting and integrating more advanced techniques to achieve better portability and adaptation capability.
	
	\section*{Acknowledgments}
	This work was partly supported by the National Key Research and Development Program of China with ID 2018AAA0103203.
	
%	\section*{References}
	
	\bibliography{manuscript}

\begin{thebibliography}{10}
\expandafter\ifx\csname url\endcsname\relax
  \def\url#1{\texttt{#1}}\fi
\expandafter\ifx\csname urlprefix\endcsname\relax\def\urlprefix{URL }\fi
\expandafter\ifx\csname href\endcsname\relax
  \def\href#1#2{#2} \def\path#1{#1}\fi

\bibitem{tradm1}
S.~Zhu, S.~A. Dianat, L.~K. Mestha, End-to-end system of license plate
  localization and recognition, Journal of Electronic Imaging 24~(2) (2015) 1
  -- 18.

\bibitem{tradm2}
Y.~Wen, Y.~Lu, J.~Yan, Z.~Zhou, K.~M. von Deneen, P.~Shi, An algorithm for
  license plate recognition applied to intelligent transportation system, IEEE
  Transactions on Intelligent Transportation Systems 12~(3) (2011) 830--845.

\bibitem{alpr}
S.~Du, M.~Ibrahim, M.~Shehata, W.~Badawy, Automatic license plate recognition
  (alpr): A state-of-the-art review, IEEE Transactions on Circuits and Systems
  for Video Technology 23~(2) (2012) 311--325.

\bibitem{pvw}
W.~Zhou, H.~Li, Y.~Lu, Q.~Tian, Principal visual word discovery for automatic
  license plate detection, IEEE Transactions on Image Processing 21~(9) (2012)
  4269--4279.

\bibitem{vlpd}
S.-L. Chen, C.~Yang, J.-W. Ma, F.~Chen, X.-C. Yin, Simultaneous end-to-end
  vehicle and license plate detection with multi-branch attention neural
  network, IEEE Transactions on Intelligent Transportation Systems 21~(9)
  (2020) 3686--3695.

\bibitem{lprec1}
S.~C. Barreto, J.~A. Lambert, F.~de~Barros~Vidal, Using synthetic images for
  deep learning recognition process on automatic license plate recognition, in:
  Mexican Conference on Pattern Recognition (MCPR), 2019, pp. 115--126.

\bibitem{reid}
J.~{\v{S}}pa{\v{n}}hel, J.~Sochor, R.~Jur{\'a}nek, A.~Herout,
  L.~Mar{\v{s}}{\'\i}k, P.~Zem{\v{c}}{\'\i}k, Holistic recognition of low
  quality license plates by cnn using track annotated data, in: 14th IEEE
  International Conference on Advanced Video and Signal Based Surveillance
  (AVSS), 2017, pp. 1--6.

\bibitem{caltech}
Caltech, Caltech license plate dataset,
  \url{http://www.vision.caltech.edu/html-files/archive.html} (2005).

\bibitem{zemris}
Zemris, Zemris license plate dataset,
  \url{http://www.zemris.fer.hr/projects/LicensePlates/hrvatski/rezultati.shtml}
  (2003).

\bibitem{ccpd}
Z.~Xu, W.~Yang, A.~Meng, N.~Lu, H.~Huang, C.~Ying, L.~Huang, Towards end-to-end
  license plate detection and recognition: A large dataset and baseline, in:
  Proceedings of the European Conference on Computer Vision (ECCV), 2018, pp.
  255--271.

\bibitem{aolp}
G.-S. Hsu, J.-C. Chen, Y.-Z. Chung, Application-oriented license plate
  recognition, IEEE Transactions on Vehicular Technology 62~(2) (2012)
  552--561.

\bibitem{ssig}
G.~R. Gon{\c{c}}alves, S.~P.~G. da~Silva, D.~Menotti, W.~R. Schwartz, Benchmark
  for license plate character segmentation, Journal of Electronic Imaging
  25~(5) (2016) 053034.

\bibitem{ufpr}
R.~Laroca, E.~Severo, L.~A. Zanlorensi, L.~S. Oliveira, G.~R. Gon{\c{c}}alves,
  W.~R. Schwartz, D.~Menotti, A robust real-time automatic license plate
  recognition based on the yolo detector, in: 2018 International Joint
  Conference on Neural Networks (IJCNN), 2018, pp. 1--10.

\bibitem{pku}
Y.~Yuan, W.~Zou, Y.~Zhao, X.~Wang, X.~Hu, N.~Komodakis, A robust and efficient
  approach to license plate detection, IEEE Transactions on Image Processing
  26~(3) (2017) 1102--1114.

\bibitem{clpd}
L.~Zhang, P.~Wang, H.~Li, Z.~Li, C.~Shen, Y.~Zhang, A robust attentional
  framework for license plate recognition in the wild, IEEE Transactions on
  Intelligent Transportation Systems 22~(11) (2021) 6967--6976.

\bibitem{stela}
L.~Deng, Y.~Gong, X.~Lu, Y.~Lin, Z.~Ma, M.~Xie, Stela: A real-time scene text
  detector with learned anchor, IEEE Access 7 (2019) 153400--153407.

\bibitem{fots}
X.~Liu, D.~Liang, S.~Yan, D.~Chen, Y.~Qiao, J.~Yan, Fots: Fast oriented text
  spotting with a unified network, in: Proceedings of the IEEE Conference on
  Computer Vision and Pattern Recognition (CVPR), 2018, pp. 5676--5685.

\bibitem{vlpdr}
C.~Zhang, Q.~Wang, X.~Li, V-lpdr: Towards a unified framework for license plate
  detection, tracking, and recognition in real-world traffic videos,
  Neurocomputing 449 (2021) 189--206.

\bibitem{cnnlpdr}
S.~M. Silva, C.~R. Jung, Real-time license plate detection and recognition
  using deep convolutional neural networks, Journal of Visual Communication and
  Image Representation 71 (2020) 102773.

\bibitem{twostage}
Y.~Kessentini, M.~D. Besbes, S.~Ammar, A.~Chabbouh, A two-stage deep neural
  network for multi-norm license plate detection and recognition, Expert
  Systems with Applications 136 (2019) 159--170.

\bibitem{mtlpr}
W.~Wang, J.~Yang, M.~Chen, P.~Wang, A light cnn for end-to-end car license
  plates detection and recognition, IEEE Access 7 (2019) 173875--173883.

\bibitem{yoloalpr}
R.~Laroca, L.~A. Zanlorensi, G.~R. Gon{\c{c}}alves, E.~Todt, W.~R. Schwartz,
  D.~Menotti, An efficient and layout-independent automatic license plate
  recognition system based on the yolo detector, IET Intelligent Transport
  Systems 15~(4) (2021) 483--503.

\bibitem{eulpr}
S.~Qin, S.~Liu, Efficient and unified license plate recognition via lightweight
  deep neural network, IET Image Processing 14~(16) (2020) 4102--4109.

\bibitem{easypr}
R.~Liu, M.~Li, Easypr, \url{https://github.com/liuruoze/EasyPR} (2014).

\bibitem{retinanet}
T.-Y. Lin, P.~Goyal, R.~Girshick, K.~He, P.~Doll{\'a}r, Focal loss for dense
  object detection, in: Proceedings of the IEEE International Conference on
  Computer Vision (ICCV), 2017, pp. 2980--2988.

\bibitem{fpn}
T.-Y. Lin, P.~Doll{\'a}r, R.~Girshick, K.~He, B.~Hariharan, S.~Belongie,
  Feature pyramid networks for object detection, in: Proceedings of the IEEE
  Conference on Computer Vision and Pattern Recognition (CVPR), 2017, pp.
  2117--2125.

\bibitem{fastrcnn}
R.~Girshick, Fast r-cnn, in: Proceedings of the IEEE International Conference
  on Computer Vision (ICCV), 2015, pp. 1440--1448.

\bibitem{crnn}
B.~Shi, X.~Bai, C.~Yao, An end-to-end trainable neural network for image-based
  sequence recognition and its application to scene text recognition, IEEE
  Transactions on Pattern Analysis and Machine Intelligence 39~(11) (2016)
  2298--2304.

\bibitem{drnn}
L.~{Deng}, Y.~{Gong}, X.~{Lu}, X.~{Yi}, Z.~{Ma}, M.~{Xie}, Focus-enhanced scene
  text recognition with deformable convolutions, in: IEEE 5th International
  Conference on Computer and Communications (ICCC), 2019, pp. 1685--1689.

\bibitem{lstm}
S.~Hochreiter, J.~Schmidhuber, Long short-term memory, Neural Computation 9~(8)
  (1997) 1735--1780.

\bibitem{ctc}
A.~Graves, S.~Fern{\'a}ndez, F.~Gomez, J.~Schmidhuber, Connectionist temporal
  classification: Labelling unsegmented sequence data with recurrent neural
  networks, in: Proceedings of the 23rd International Conference on Machine
  Learning (ICML), 2006, pp. 369--376.

\bibitem{pytorch}
A.~Paszke, S.~Gross, F.~Massa, A.~Lerer, J.~Bradbury, G.~Chanan, T.~Killeen,
  Z.~Lin, N.~Gimelshein, L.~Antiga, et~al., Pytorch: An imperative style,
  high-performance deep learning library, Advances in Neural Information
  Processing Systems (NIPS) 32 (2019) 8026--8037.

\bibitem{maskrcnn}
K.~He, G.~Gkioxari, P.~Doll{\'a}r, R.~Girshick, Mask r-cnn, in: Proceedings of
  the IEEE International Conference on Computer Vision (ICCV), 2017, pp.
  2961--2969.

\bibitem{ssd}
W.~Liu, D.~Anguelov, D.~Erhan, C.~Szegedy, S.~Reed, C.-Y. Fu, A.~C. Berg, Ssd:
  Single shot multibox detector, in: European Conference on Computer Vision
  (ECCV), 2016, pp. 21--37.

\bibitem{yolov3}
J.~Redmon, A.~Farhadi, Yolov3: An incremental improvement, arXiv preprint
  arXiv:1804.02767.

\bibitem{yolov4}
A.~Bochkovskiy, C.-Y. Wang, H.-Y.~M. Liao, Yolov4: Optimal speed and accuracy
  of object detection, arXiv preprint arXiv:2004.10934.

\bibitem{scaledyolov4}
C.-Y. Wang, A.~Bochkovskiy, H.-Y.~M. Liao, Scaled-yolov4: Scaling cross stage
  partial network, in: Proceedings of the IEEE/CVF Conference on Computer
  Vision and Pattern Recognition (CVPR), 2021, pp. 13029--13038.

\bibitem{fasterrcnn}
S.~Ren, K.~He, R.~Girshick, J.~Sun, Faster r-cnn: Towards real-time object
  detection with region proposal networks, in: Advances in Neural Information
  Processing Systems (NIPS), 2015, pp. 91--99.

\bibitem{wang2007cascade}
S.-Z. Wang, H.-J. Lee, A cascade framework for a real-time statistical plate
  recognition system, IEEE Transactions on Information Forensics and Security
  2~(2) (2007) 267--282.

\bibitem{yolov2}
J.~Redmon, A.~Farhadi, Yolo9000: Better, faster, stronger, in: Proceedings of
  the IEEE Conference on Computer Vision and Pattern Recognition (CVPR), 2017,
  pp. 7263--7271.

\bibitem{te2e}
H.~Li, P.~Wang, C.~Shen, Toward end-to-end car license plate detection and
  recognition with deep neural networks, IEEE Transactions on Intelligent
  Transportation Systems 20~(3) (2019) 1126--1136.

\end{thebibliography}
	
\end{document}